\documentclass{article}

\PassOptionsToPackage{numbers, compress}{natbib}
\usepackage[final]{timeseries_workshop}

\usepackage[utf8]{inputenc} % allow utf-8 input
\usepackage[T1]{fontenc}    % use 8-bit T1 fonts
\usepackage{hyperref}       % hyperlinks
\usepackage{url}            % simple URL typesetting
\usepackage{booktabs}       % professional-quality tables
\usepackage{amsfonts}       % blackboard math symbols
\usepackage{nicefrac}       % compact symbols for 1/2, etc.
\usepackage{microtype}      % microtypography
\usepackage{xcolor}         % colors

\usepackage{natbib}

\hypersetup{
    colorlinks=true,
    linkcolor=blue,
    filecolor=magenta,      
    urlcolor=blue,
    citecolor=blue,
}

\newcommand{\model}[1]{\texttt{#1}}
\newcommand{\MLP}{\model{MLP}}
\newcommand{\DLinear}{\model{DLinear}}
\newcommand{\TCN}{\model{TCN}}
\newcommand{\LSTM}{\model{LSTM}}
\newcommand{\RNN}{\model{RNN}}
\newcommand{\TimesNet}{\model{TimesNet}}
\newcommand{\TSMixer}{\model{TSMixer}}

\newcommand{\NHITS}{\model{NHITS}}

\newcommand{\VanillaTransformer}{\model{VanillaTransformer}}
\newcommand{\TFT}{\model{TFT}}
\newcommand{\PatchTST}{\model{PatchTST}}

\newcommand{\Autoformer}{\model{Autoformer}}
\newcommand{\Informer}{\model{Informer}}
\newcommand{\iTransformer}{\model{iTransformer}}

\newcommand{\TimesFM}{\model{TimesFM}}
\newcommand{\MOMENT}{\model{MOMENT}}
\newcommand{\LagLlama}{\model{LagLlama}}
\newcommand{\Moirai}{\model{Moirai}}
\newcommand{\TimeGPT}{\model{TimeGPT}}
\newcommand{\Chronos}{\model{Chronos}}
\newcommand{\Timer}{\model{Timer}}
\newcommand{\TinyTimeMixers}{\model{Tiny Time Mixers}}

\usepackage{caption}
\usepackage{subcaption}
\usepackage{multirow}
\usepackage{graphicx} 
\usepackage{amsmath}
\usepackage{xcolor}
\usepackage{todonotes}

\title{Implicit Reasoning in Deep Time Series Forecasting}

\author{Willa Potosnak$\textsuperscript{1}$, Cristian Challu$\textsuperscript{1, 2}$, Mononito Goswami$\textsuperscript{1}$, Michał Wiliński$\textsuperscript{1}$, \\
\textbf{Nina Żukowska$\textsuperscript{1}$}, \textbf{Artur Dubrawski$\textsuperscript{1}$} \\
	$\textsuperscript{1}$Auton Lab, School of Computer Science, Carnegie Mellon University \\
	$\textsuperscript{2}$Nixtla \\
}

\begin{document}

\thanks{Correspondence to: Willa Potosnak <wpotosna@andrew.cmu.edu>}

\maketitle

\begin{abstract}
  Recently, time series foundation models have shown promising zero-shot forecasting performance on time series from a wide range of domains. However, it remains unclear whether their success stems from a true understanding of temporal dynamics or simply from memorizing the training data. While implicit reasoning in language models has been studied, similar evaluations for time series models have been largely unexplored. This work takes an initial step toward assessing the reasoning abilities of deep time series forecasting models. We find that certain linear, MLP-based, and patch-based Transformer models generalize effectively in systematically orchestrated out-of-distribution scenarios, suggesting underexplored reasoning capabilities beyond simple pattern memorization.
\end{abstract}

% \todo[color=red!40, inline]{Willa to-dos: 1.) fix data section, 2.) Add plots of subtraction experiment embedding as ratio of aggregates to components increases (to appendix), 3.) get patchdecoder and patchencoderdecoder results}

% \todo[color=red!40, inline]{Questions: 
% 1.) Should we include our MAE frequency vs. trend variation scatter plots in the appendix? 2.) should I add other forecast example plots to the appendix? like those for sinusoidal function composition 3.) I would very much appreciate feedback on the methods section. I've tried to align the notation with that in language model papers as best as possible).
% }

\section{Introduction}
Foundation models have demonstrated an exceptional ability to generalize to previously unseen data in zero-shot prediction tasks. Inspired by the success of such models in Natural Language Processing, recent work has adapted Transformers to build time series foundation models (TSFM). Zero-shot inference is particularly important for time series models, which must handle complex patterns, seasonal variations, and emerging trends where little to no reference data may be available. 

%Foundation models are typically trained on large, diverse datasets, but this raises a critical question in the context of time series forecasting:
Foundation models are trained on large, diverse datasets, raising a critical question for time series forecasting: \textbf{do these models generalize well because they learn underlying concepts of temporal dynamics, or do they simply memorize specific patterns seen during training?} If models rely on memorization, particularly in the form of time series pattern matching, it could lead to redundant knowledge storage, parameter inefficiency, and limit their ability to generalize well to out-of-distribution (OOD) data. Ideally, a TSFM should be capable of implicit reasoning, allowing it not to depend solely on memorization but to infer latent temporal dynamics. Such models would be able to generalize from fewer data points, offering enhanced parameter efficiency and robustness.

While extensive research has been conducted to evaluate memorization and implicit reasoning in language models, similar evaluations for time series models have been largely unexplored. In this work, we take an initial step toward evaluating the implicit reasoning capabilities of time series models in forecasting tasks. Our findings highlight the potential of linear, MLP-based, and patch-based Transformer models to perform well in carefully orchestrated OOD scenarios, suggesting that these models may have untapped capabilities in reasoning beyond mere memorization.
 \label{section:introduction}
\section{Related work}
\paragraph{Implicit Reasoning in LLMs.} Prior research has explored implicit reasoning in language models \citep{allenshu2023physics3.2, allenzhu2023physics3.1, wang2024grokked, soheeyang2024multihopreasoning,  zhong2023mquake}. Implicit reasoning is often assessed through tasks that require models to apply knowledge learned during training to new test instances. One common form is \textit{composition}, where models must chain multiple facts to answer a question \citep{wang2024grokked, soheeyang2024multihopreasoning, zhong2023mquake}. Other forms of implicit reasoning explored include \textit{comparison} and \textit{inverse search} \citep{allenshu2023physics3.2, wang2024grokked}. Comparison involves models evaluating two or more entities to make judgments, such as determining whether the attribute value of one entity is greater or smaller than that of another. Inverse search tests a model's ability to generate predictions in the reverse order of the training task. For example, this could involve applying the model to identify an entity based on its attributes when it was originally trained to predict the attributes of entities. More information on related work is provided in Appendix~\ref{apd:extended_related_work}. No prior research has conducted controlled experiments in time series forecasting to evaluate implicit reasoning on OOD data. Our study addresses this gap by introducing a novel framework that aligns LLM research with time series models, offering insights into optimal architectures for future TSFM development.

%This provides comparative insights into implicit reasoning, aiding in identifying optimal architectures for future TSFM development.

\paragraph{Time Series Foundation Models.} Several foundation models, including \Chronos\ \citep{ansari2024chronos}, \LagLlama~\citep{rasul2024lagllama}, \Moirai~\citep{moirai2024}, \MOMENT\ \citep{goswami2024moment}, \TimesFM\ \citep{das2024TimesFM}, \TimeGPT~\citep{garza2023timegpt1}, \Timer~\citep{liutimer}, and \TinyTimeMixers~\cite{ekambaram2024ttms}, have been developed for time series forecasting. Some studies have also examined the impact of learning with synthetic data \citep{ansari2024chronos, das2024TimesFM}. \MOMENT\ analyzed embeddings from synthetic sinusoids to isolate components like trends, while \Chronos\ showed strong performance in forecasting composite series but struggled with exponential trends. These works demonstrate that TSFMs can learn distinct time series functions, though it remains unclear if their success is due to large-scale training or inherent reasoning abilities.

 \label{section:related_work}
\section{Methods}
% composition --> functional composition and time series composition are different things (for example use signal composition)

% f1 \in F^{(in)}_{sc}   f2 \in F^{(out)}_{sc}   with F^{(in)}_{sc}, F^{(out)}_{sc} \subset F_{sc} and F^{(in)}_{sc} \intersect F^{(out)}_{sc}
%(x,f1, y) \in D^{in} (x,f2, y) \in D^{out}. Talk about restricting the parameter space in the table.

\subsection{Implicit Reasoning Tasks}
We assess the implicit reasoning capabilities of models, or their ability to apply and manipulate internalized patterns in novel situations, through experiments inspired by language model research \cite{allenshu2023physics3.2, allenzhu2023physics3.1, wang2024grokked, soheeyang2024multihopreasoning, zhong2023mquake}. We carefully design experiments to test 3 forms of implicit reasoning: \textit{composition}, \textit{comparison}, and \textit{inverse search}. We follow the notation from \citep{wang2024grokked, zhong2023mquake}, where `facts' are formatted as \textit{(subject, relation, object)}. For continuous data, we adapt this to \textit{(input, function, output)}, with $f(x) = y$ defining the relationship between the function $f$, input $x$, and output $y$. We consider two sets of basis functions: $\mathcal{F}_c = \{f:f(x) = mx \mid m \in \mathbb{R}\}$ and $\mathcal{F}_s = \{f:f(x) = a\sin(b 2\pi x) + c \mid a, b, c \in \mathbb{R}\}$ where $\mathcal{F}_c$ is the set of all constant signals with linear trends $m$ and $\mathcal{F}_s$ is the set of all sinusoidal signals with amplitude $a$, frequency $b$, and baseline shift $c$. We also define a set of functions consisting of compositions of basis functions: $\mathcal{F}_{sc} = \{f:f(x) = a\sin(b 2\pi x) + c \, \{\pm, \times\} \, mx \mid a, b, c, m \in \mathbb{R}\}$. For a set of data-generating functions and their parameters included in the training set and referred to as in-distribution (ID) or $\mathcal{P}_{\text{train}}$, we assess a model's ability to generalize to other disjoint sets of data-generating functions observed only at inference, denoted as $\mathcal{P}_{\text{test}}$, which are considered OOD.

\begin{figure}[t]
     \centering
     \begin{subfigure}[b]{0.325\textwidth}
         \centering
         \includegraphics[width=\textwidth]{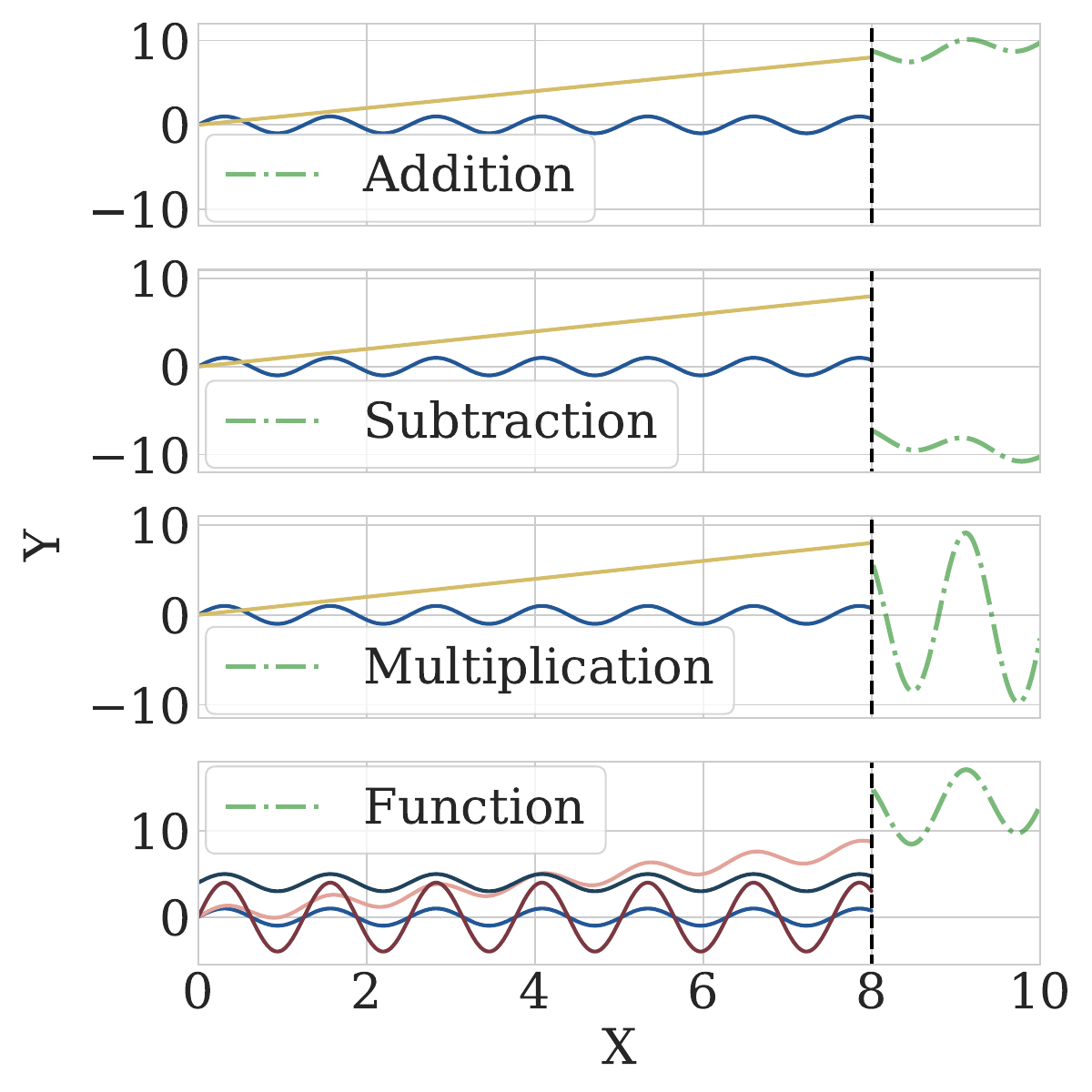}
         \caption{Composition}
         \label{fig:compositon_examples}
     \end{subfigure}
     \begin{subfigure}[b]{0.325\textwidth}
         \centering
         \includegraphics[width=\textwidth]{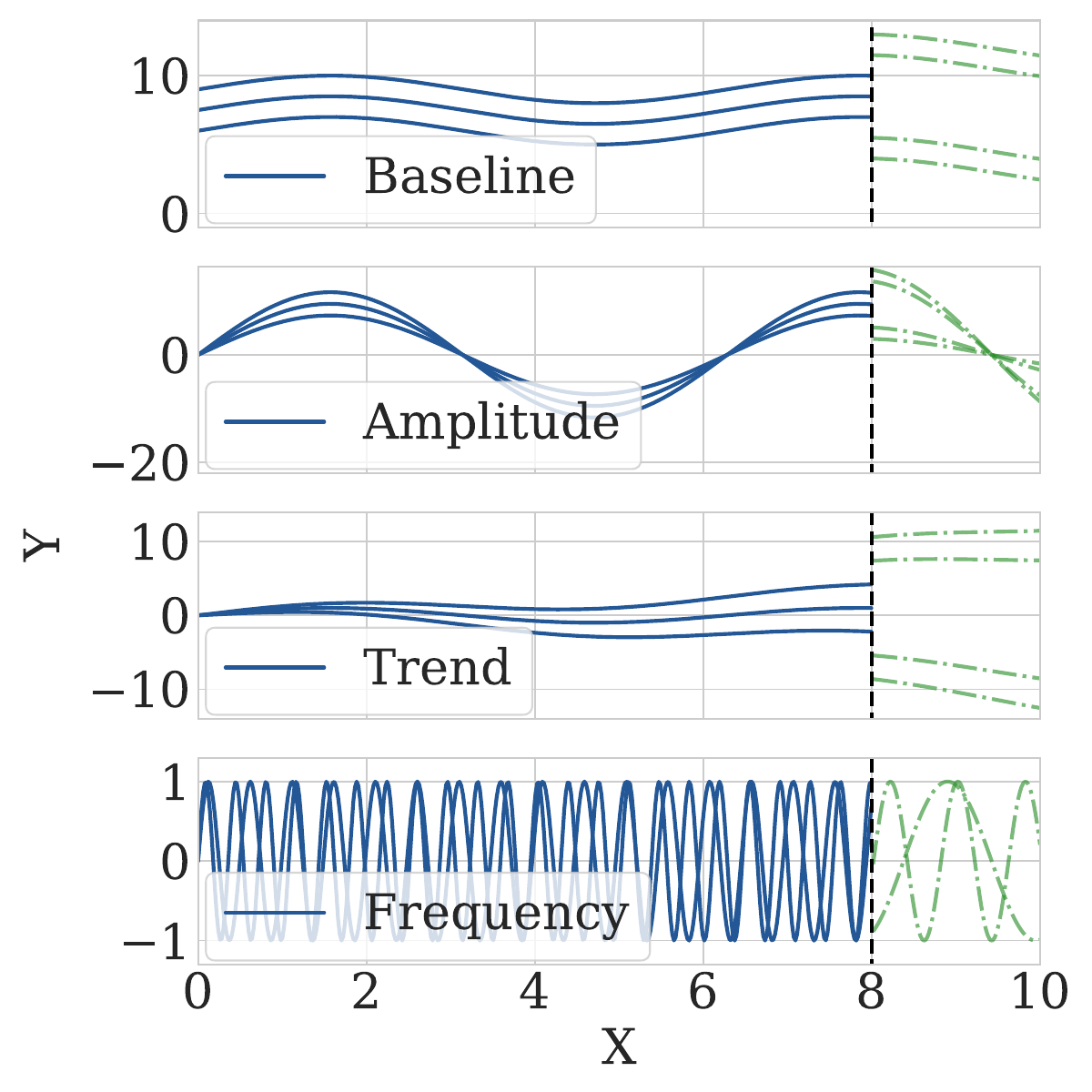}
         \caption{Comparison}
         \label{fig:comparison_examples}
     \end{subfigure}
     \begin{subfigure}[b]{0.325\textwidth}
         \centering
         \includegraphics[width=\textwidth]{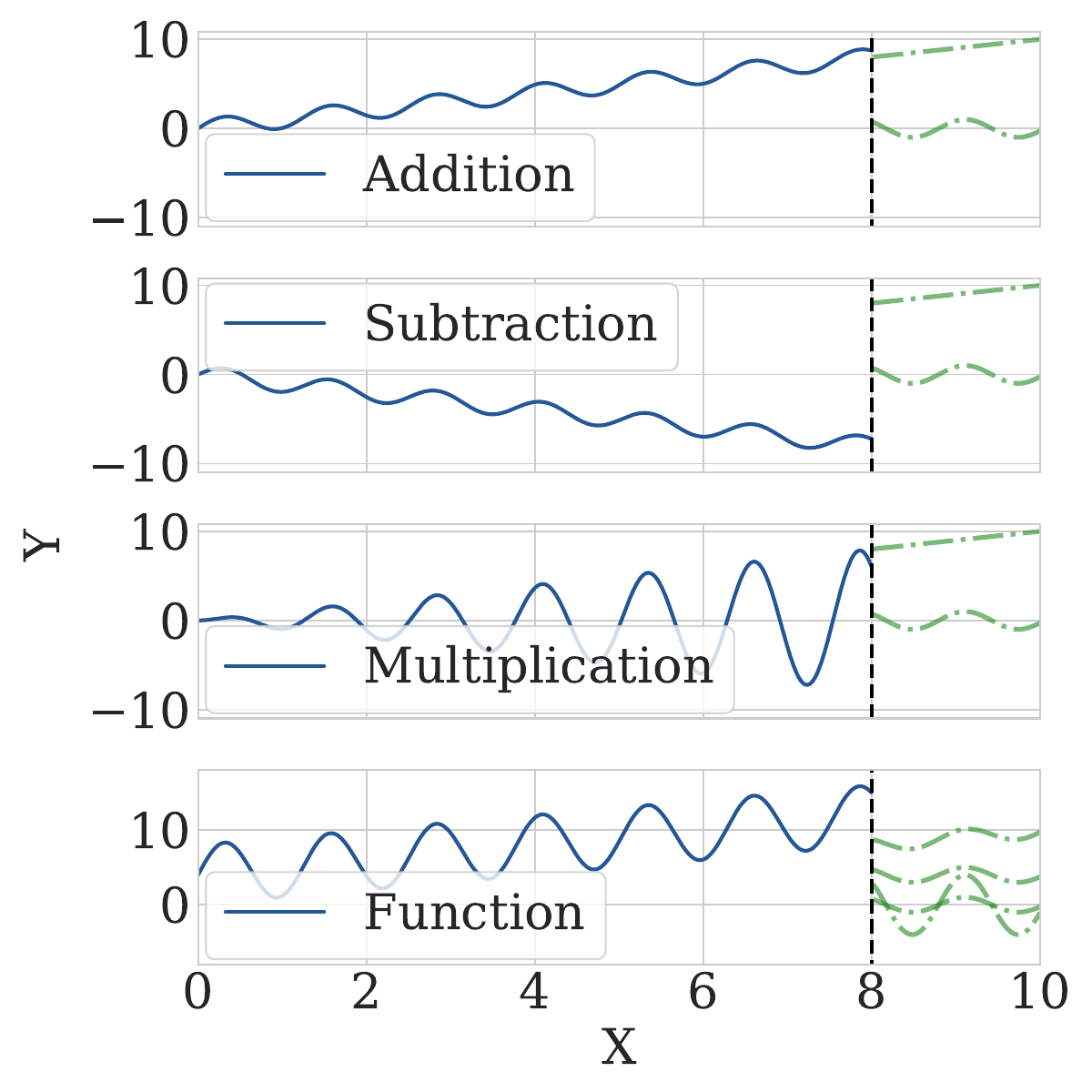}
         \caption{Inverse Search}
         \label{inverse_search_examples}
     \end{subfigure}
        % \caption{Implicit reasoning tasks for time series include: composition (a), comparison (b), and inverse search (c). In composition tasks, models trained on time series components are evaluated on their ability to predict new composite series seen only during inference. In comparison tasks, models trained on time series are evaluated on their ability to forecast series derived with function values not observed during training. In inverse search tasks, models trained on composite time series are evaluated on their ability to forecast decomposed component series.
        \caption{Implicit reasoning tasks for time series include composition (a), comparison (b), and inverse search (c). In composition, models predict new composite series seen only at inference. In comparison, they forecast series from unseen function values. In inverse search, models predict decomposed component series. The black dashed line in the figure indicates in-distribution time series observed during training (left) and out-of-distribution data observed during inference (right).}
        \label{fig:implicit_reasoning_examples}
\end{figure}

\paragraph{Composition.}
For models trained on signals generated from basis functions considered ID, we assess whether they can accurately forecast signal compositions that are considered OOD. Specifically, we define the datasets as follows: $\mathcal{P}_{\text{train}}=\mathcal{F}_s \cup \mathcal{F}_c \text{ and } \mathcal{P}_{\text{test}}=\mathcal{F}_{sc}, \text{ where } \mathcal{P}_{\text{train}}\cap \mathcal{P}_{\text{test}} = \emptyset$. Consider a specific example:
\begin{align}
f_s\in \mathcal{F}_s, f_c \in \mathcal{F}_c, f_{sc} \in \mathcal{F}_{sc}, \forall x \in \mathbb{R}, (x, f_s, y), (x, f_c, y)\in \mathcal{P}_{\text{train}} \text{ and } (x, f_{sc}, y)\in \mathcal{P}_{\text{test}}
\end{align}
%\begin{align}
% f_1, f_2 \in \mathcal{F}, \, \forall x \in \mathbb{R}, (x, f_1, \widecheck{x}) \land (\widecheck{x}, f_2, y) \Rightarrow f_2(f_1(x)) = y 
% \end{align}
Models are trained exclusively on the individual component series, and the compositions of series are withheld for final evaluations as illustrated in Fig.~\ref{fig:compositon_examples}. We train models on 30 trend and seasonality series ($n=60$) and evaluate them on all compositions of these series ($n=900$).

\paragraph{Comparison.}
Comparison refers to a model’s ability to make comparative judgments about the attributes or parameter values of facts \citep{wang2024grokked}. For models trained on data generated from functions considered ID, we evaluate whether these models can generalize to the same data-generating function with larger or smaller parameter values that are considered OOD. Specifically, we define the datasets as follows: $\mathcal{P}_{\text{train}}, \mathcal{P}_{\text{test}} \subset \mathcal{F}_{sc}, \text{ where } \mathcal{P}_{\text{train}} \cap \mathcal{P}_{\text{test}} = \emptyset$. Consider a specific example:
\begin{align}
f_1 \in \mathcal{P}_{\text{train}}, f_2 \in \mathcal{P}_{\text{test}}, \forall x \in \mathbb{R}, (x,f_1, y) \in \mathcal{P}_{\text{train}} \text{ and } (x,f_2, y) \in \mathcal{P}_{\text{test}}.
\end{align}
% \begin{align}
% f_1, f_2 \in \mathcal{F}, \, \forall x \in \mathbb{R}, (x, f_1, \widecheck{x}) \land (\widecheck{x}, f_2, y) \land f_1(x) < y \Rightarrow f_1(x) < f_2(f_1(x)) \\
% f_1, f_2 \in \mathcal{F}, \, \forall x \in \mathbb{R}, (x, f_1, \widecheck{x}) \land (\widecheck{x}, f_2, y) \land f_1(x) > y \Rightarrow  f_1(x) > f_2(f_1(x))
% \end{align}
We train models on signals with one varying parameter: $a, b, c, \text{ or } m$ ($n=1200$) and evaluate them on signals with smaller or larger parameter values not seen during training ($n=120$).

\paragraph{Inverse search.}
We evaluate whether models can decompose functions through the inverse search task. For models trained on signal compositions, we assess their ability to accurately forecast individual basis functions comprising the signal, which are considered OOD. We define the datasets as follows: $\mathcal{P}_{\text{train}}=\mathcal{F}_{sc} \text{ and }  \mathcal{P}_{\text{test}}=\mathcal{F}_s \cup \mathcal{F}_c, \text{ where } \mathcal{P}_{\text{train}}\cap \mathcal{P}_{\text{test}} = \emptyset$. Consider a specific example:
\begin{align}
f_s\in \mathcal{F}_s, f_c \in \mathcal{F}_c, f_{sc} \in \mathcal{F}_{sc}, \forall x \in \mathbb{R}, (x, f_{sc}, y)\in\mathcal{P}_{\text{train}} \text{ and } (x, f_s, y), (x, f_c, y)\in \mathcal{P}_{\text{test}}
\end{align}
% \begin{align}
% f_1, f_2 \in \mathcal{F}, \, \forall x \in \mathbb{R}, (x, f_1, y) \land (y, f_2, \widecheck{x}) \Rightarrow f_2(f_1(x)) = \widecheck{x} \land \forall f_i \in f_1, \; f_i(x_i) = y_i
% \end{align}
We train models on composite addition of trend and seasonality series ($n=900$). We then evaluate the models on all decomposed trend and seasonality components series ($n=60$).

\subsection{Data}
We use the functions in Table~\ref{tab:synthetic_data_parameters}, with parameter values detailed in Appendix~\ref{apd:synthetic_data_parameters} to generate synthetic data. Both ID and OOD data use $x\in[0,1]$ with 1200 samples. Models are trained on the first 1000 ID samples and evaluated on the last 200 OOD samples, testing their implicit reasoning capabilities.

% \begin{table}[h]
%     \centering
%     \caption{Synthetic dataset functions and parameters for implicit reasoning tasks.}
%     \label{tab:hyperparameters}
%     \begin{tabular}{r|cll} 
%     \toprule
%     \textbf{Task} & \textbf{Set} & \textbf{Function} & \textbf{Parameters}\\
%     \hline
%     \textbf{Composition} & ID & $f_1(x)=\{\sin(B2\pi x), Mx\}$ & \multirow{4}{*}{$B, M$} \\
%     (Addition) & OOD & $f_2(x) = \sin(B2\pi x)\!+\!Mx$ & {} \\
%     (Subtraction) & OOD & $f_2(x) = \sin(B2\pi x)\!-\!Mx$ & {} \\
%     (Multiplication) & OOD & $f_2(x) = \sin(B2\pi x)\!\times\!Mx$ & {} \\
%     \hline
%     \multirow{2}{*}{\textbf{Composition}} & \multirow{2}{*}{ID} & $f_1(x)=\{A\sin(2\pi x), \sin(B2\pi x), $ & \multirow{3}{*}{$A, B, C, M$} \\
%     \multirow{2}{*}{(Function)} & {} & \hspace{3em}$\sin(2\pi x)\!+\!C, \sin(2\pi x)\!+\!Mx\}$ \\
%     {} & OOD & $f_2(x)=A\sin(B2\pi x)\!+\!C\!+\!Mx$ & {}\\
%     \hline
%     \multirow{3}{*}{\textbf{Comparison}} & ID & $f_1(x)=A\sin(B2\pi x)\!+\!D\!+\!Mx$ & $A, B, C, M$ \\
%     {} & \multirow{2}{*}{OOD} & \multirow{2}{*}{$f_2(x)=a\sin(b2\pi x)\!+\!c\!+\!mx$} & $a\!<\!A\!\lor\!a\!>\!A$, $b\!<\!B\!\lor\!b\!>\!B$ \\
%     {} & {} & {} & $c\!<\!C\!\lor\!c\!>\!C$, $c\!<\!C\!\lor\!c\!>\!C$ \\
%     \hline
%     \multirow{2}{*}{\textbf{Inverse Search}} & ID & $f_2(x) = \sin(B2\pi x)\!+\!Mx$ & \multirow{2}{*}{$B, M$} \\
%     {} & OOD & $f_1(x)=\{\sin(B2\pi x), Mx\}$ & {} \\
%     \bottomrule
%     \end{tabular}
%   \label{tab:synthetic_data_parameters}
%   \centering
% \end{table}

\begin{table}[t]
    \caption{Synthetic dataset functions and parameters for implicit reasoning tasks.}
    \label{tab:hyperparameters}
    \centering
    \resizebox{0.9\textwidth}{!}{
    \begin{tabular}{l|lll} 
    \toprule
    \textbf{Task} & \textbf{Dataset} & \textbf{Function} & \textbf{Parameters}\\
    \hline
    \hline
    \textbf{Composition} & $\mathcal{P}_{\text{train}}$ & $f_s\in \mathcal{F}_s, f_c \in \mathcal{F}_c$ & \multirow{4}{*}{$a\!=\!0 \land b \!\in\! [b_l, b_u] \land c\!=\!0 \land m \!\in\! [m_l, m_u]$} \\
    Addition & $\mathcal{P}_{\text{test}}$ & $f_s\!+\!f_c \in \mathcal{F}_{sc}$ & {} \\
    Subtraction & $\mathcal{P}_{\text{test}}$ & $f_s\!-\!f_c \in \mathcal{F}_{sc}$ & {} \\
    Multiplication & $\mathcal{P}_{\text{test}}$ & $f_s\!\times\!f_c \in \mathcal{F}_{sc}$ & {} \\
    \hline
    \textbf{Composition} & $\mathcal{P}_{\text{train}}$ & $f_s\!+\!f_c \in \mathcal{F}_{sc}$ & $a \!\in\! [a_l, a_u] \lor b \!\in\! [b_l, b_u] \lor c \!\in\! [c_l, c_u] \lor m \!\in\! [m_l, m_u]$  \\
    Function & $\mathcal{P}_{\text{test}}$ & $f_s\!+\!f_c \in \mathcal{F}_{sc}$ & $a \!\in\! [a_l, a_u] \land b \!\in\! [b_l, b_u] \land c \!\in\! [c_l, c_u] \land m \!\in\! [m_l, m_u]$  \\
    \hline
    \hline
    \multirow{2}{*}{\textbf{Comparison}} & $\mathcal{P}_{\text{train}}$ & $f_s\!+\!f_c \in \mathcal{F}_{sc}$ & $a \!\in\! [a_{l}, a_{u}] \lor b \!\in\! [b_{l}, b_{u}] \lor c \!\in\! [c_{l}, c_{u}] \lor m \!\in\! [m_{l}, m_{u}]$ \\
    {} & $\mathcal{P}_{\text{test}}$ & $f_s\!+\!f_c \in \mathcal{F}_{sc}$ & $i \!<\! i_l \lor i\!>i_u \text{ for } i \!=\! a \lor i \!=\! b \lor i \!=\! c \lor i \!=\! m$ \\
    \hline
    \hline
    \textbf{Inverse} & $\mathcal{P}_{\text{train}}$ & $f_s\!+\!f_c \in \mathcal{F}_{sc}$ & \multirow{2}{*}{$a\!=\!0 \land b \!\in\! [b_l, b_u] \land c\!=\!0 \land m \!\in\! [m_l, m_u]$} \\
    \textbf{Search} & $\mathcal{P}_{\text{test}}$ & $f_s \in \mathcal{F}_s, f_c \in \mathcal{F}_c$ & {} \\
    \bottomrule
    \end{tabular}}
  \label{tab:synthetic_data_parameters}
  \centering
\end{table}

\subsection{Models}
We trained models using 13 algorithm implementations obtained from the \texttt{Neuralforecast}~\citep{olivares2022library_neuralforecast} library. Trained models include: Multi-Layer Perceptron (\MLP)~\citep{rosenblatt1958_mlp}, \DLinear~\citep{zeng_2023_dlinear}, \NHITS~\cite{challu_olivares2022_nhits}, \TSMixer~\citep{chen2023tsmixer}, Long-Short Term Memory (\LSTM)~\citep{sak2014_lstm}, Temporal Convolution Network (\TCN)~\citep{bai2018_tcn, oord2016_tcn}, \TimesNet~\citep{wu2023timesnettemporal2dvariationmodeling}, \VanillaTransformer~\citep{vaswani_2021_attentionisallyouneed, zhou2021informerefficienttransformerlong}, Temporal Fusion Transformer (\TFT)~\citep{lim2021_tft}, \Autoformer~\citep{wu_2021_autoformer}, \Informer~\citep{zhou2021informerefficienttransformerlong}, inverted Transformer (\iTransformer)~\citep{liu2024itransformerinvertedtransformerseffective}, and patch time series Transformer (\PatchTST)~\citep{nie2023patchtst}. This controlled setup is crucial for directly comparing architectures under identical conditions. By evaluating Transformer models with key components like attention and time-series patching, aligned to various TSFMs, our work delivers valuable insights into how TSFMs may leverage implicit reasoning in OOD generalization. We also generate zero-shot forecasting results for pretrained \Chronos, \LagLlama, \TimesFM, and \MOMENT. More information on these models, as well as model training and hyperparameters, is provided in Appendix~\ref{apd:first_models} and \ref{apd:first_hyperparameters}, respectively.

% Our code implementation is open-source and can be found on Github.
% \footnote{\textbf{\texttt{\href{}{Link to be added following paper acceptance.}}}}
 \label{section:methods}
\section{Results}
\begin{figure}[t]
     \centering
     \includegraphics[width=\textwidth]{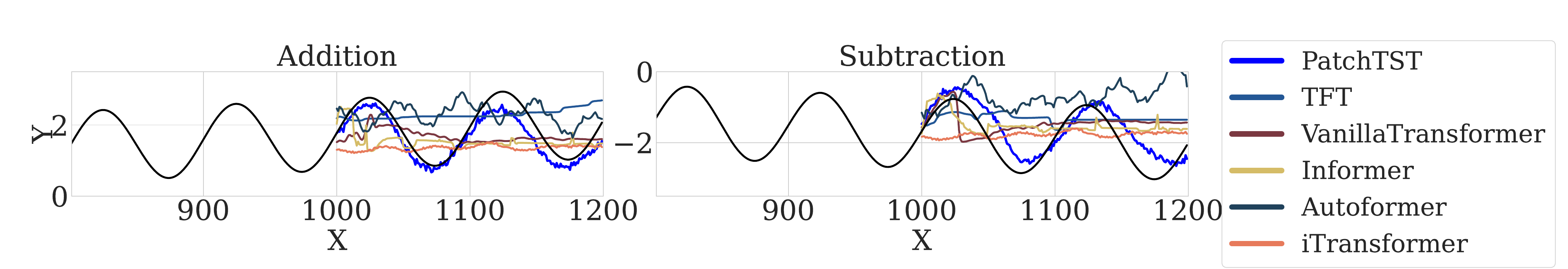}
    \caption{PatchTST shows capability in generating forecasts that accurately capture the composition of trend and seasonality series compared to other Transformer models.}
    \label{fig:main_forecasts}
\end{figure}

\PatchTST\ outperformed all other evaluated Transformer models on the composition task, particularly for addition and subtraction tasks as shown in Fig~\ref{fig:main_forecasts}. However, the model's performance on these tasks is sensitive to patch length, as shown in Fig.~\ref{fig:addition_forecast_examples} in Appendix~\ref{section:appendixB}, indicating the critical role of patch length in the model's knowledge manipulation capabilities. Table~\ref{tab:main_results_table} includes Mean Absolute Error (MAE) computed across all points within the forecast horizon and averaged across series for each task for \DLinear, \NHITS, and Transformer models. As shown in Table~\ref{tab:main_results_table}, \DLinear\ had the smallest forecasting error for addition and subtraction composition tasks as well as comparison tasks, while \NHITS\ had the smallest error for multiplication composition, function composition, and inverse search tasks. Results for each task, including standard deviations, for all 15 models are shown in Tables~\ref{tab:composition_results_table}, \ref{tab:comparison_results_table}, and \ref{tab:inverse_search_results_table} in Appendix~\ref{section:appendixB}. These tables also include baseline results for models trained on the first 1000 samples of OOD data. Example forecasts for composition and inverse search tasks are shown in Figs.~\ref{fig:compositon_forecast_examples} and \ref{fig:inverse_search_forecast_examples} in Appendix~\ref{section:appendixB}. MAE formulation is provided in~\ref{apd:first_metrics}.

\begin{table}[t]
\caption{Mean Absolute Error (MAE) computed across all forecast points within the specified time horizon and averaged across series for each task. Best results are in bold, second-best in blue.}
\label{tab:main_results_table}
\centering
\resizebox{0.9\textwidth}{!}{
\begin{tabular}{r|c c c c|c|c }
\toprule
\multirow{2}{*}{\textbf{Model}} & \multicolumn{4}{c|}{\textbf{Composition}} & \multirow{2}{*}{\textbf{Comparison}} & \multirow{2}{*}{\textbf{Inverse Search}} \\
{} & \textbf{Add.} & \textbf{Sub.} & \textbf{Mult.} & \textbf{Func.} & {} & {} \\ \midrule
\DLinear & \textbf{0.689} & \textbf{0.935} & 10.522 & 21.155 & \textbf{1.072} & 0.539 \\
\NHITS & 1.427 & 2.36 & \textbf{2.065} & \textbf{3.868} & 2.218 & \textbf{0.214} \\
\VanillaTransformer & 2.432 & 2.993 & 8.249 & 11.291 & 2.120 & 0.551 \\
\TFT & \textcolor{blue}{1.024} & 3.243 & 13.395 & 10.159 & 2.045 & 0.574 \\
\Autoformer & 1.708 & 3.999 & 12.845 & 13.383 & 3.966 & 1.180 \\
\Informer & 2.045 & 3.025 & 9.246 & 12.268 & 2.131 & 0.877 \\
\iTransformer & 1.358 & 2.994 & 15.197 & 31.632 & 2.885 & 1.164 \\
\PatchTST & 1.717 & \textcolor{blue}{2.168} & 3.497 & \textcolor{blue}{6.209} & \textcolor{blue}{1.740} & \textcolor{blue}{0.394} \\
\bottomrule
% \multirow{2}{*}{\PatchDecoder} & x & x & x & x & x & x & x & x \\
%                       & \small{(x)} & \small{(x)} & \small{(x)} & \small{(x)} & \small{(x)} & \small{(x)} & \small{(x)} & \small{(x)} \\
% \hline
% \multirow{2}{*}{\PatchEncoderDecoder} & x & x & x & x & x & x & x & x \\
%                       & \small{(x)} & \small{(x)} & \small{(x)} & \small{(x)} & \small{(x)} & \small{(x)} & \small{(x)} & \small{(x)} \\
% \hline
\end{tabular}}
\end{table}

% \begin{figure}
%      \centering
%      \begin{subfigure}[b]{0.5\textwidth}
%          \centering
%          \includegraphics[width=\textwidth]{images/embedding_forecast_subplot.pdf}
%          \caption{Composition}
%          \label{fig:transformer_forecast_comparison}
%      \end{subfigure}
%      \begin{subfigure}[b]{0.1\textwidth}
%          \centering
%          \includegraphics[width=\textwidth]{images/embeddings.pdf}
%          \caption{Comparison}
%          \label{fig:three sin x}
%      \end{subfigure}
%      \hfill
%         % \caption{We compare transformer models on out-of-distribution (OOD) addition and subtraction forecasting tasks, using models trained exclusively on trend and seasonality components. PatchTST shows strong capability in generating forecasts that accurately capture the composition of trend and seasonality series. Notably, PatchTST's embeddings reveal distinct patterns during inference, despite being trained on the same dataset of positive trend and seasonality series.}
%         \label{fig:patchtst_embeddings}
% \end{figure}
 \label{section:results}
\section{Discussion}
Our findings suggest that linear, MLP-based, and patch-based Transformer models, like \PatchTST, offer implicit reasoning capabilities beyond memorization. \PatchTST\ demonstrates strong OOD forecasting performance, supporting the use of patching input time series in TSFMs like \MOMENT, \Moirai, and \TimesFM. Moreover, \NHITS's robust performance across tasks highlights the value of signal decomposition and hierarchical architectures, which if incorporated in future TSFMs could potentially enhance their forecasting capabilities. Our evaluation also shows the need for metrics beyond traditional loss measures. For example, despite \TFT's second-best MAE in addition composition, it forecasts a straight line as shown in Fig.~\ref{fig:main_forecasts}, demonstrating the limitations of loss functions in capturing forecast pattern quality in complex tasks.

% Defining time series as in-distribution (ID) and out-of-distribution (OOD) needs to consider many factors as compare to other forms of data, such as language. For example, OOD data may be defined as an entity, such as a person name or person's attribute, not seen during training. While for time series forecasting, OOD data could encompass multiple types of deviations such as frequency misalignment, differing domains, presence of anomalies, and shifts in seasonality, among others.

 \label{section:discussion}

\section*{Acknowledgment}
This work was partially supported by the NSF (awards 2406231 and 2427948), NIH (awards R01NS124642 and R01DK131586), DARPA (HR00112420329), and the US Army (W911NF-20-D0002). We would also like to thank Kin Gutierrez Olivares for his feedback on our manuscript.

\bibliography{references}

\appendix
\newpage
\section{Supplemental Information}

\subsection{Extended Related Work}\label{apd:extended_related_work}
\paragraph{Implicit Reasoning in LLMs.}  Prior research has explored implicit reasoning in language models \citep{allenshu2023physics3.2, allenzhu2023physics3.1, wang2024grokked, soheeyang2024multihopreasoning,  zhong2023mquake}. Implicit reasoning is often assessed through tasks that require models to apply knowledge learned during training to new test instances. One common form is \textit{composition}, where models must chain multiple facts to answer a question \citep{wang2024grokked, soheeyang2024multihopreasoning, zhong2023mquake}. Other forms of implicit reasoning explored include \textit{comparison} and \textit{inverse search} \citep{allenshu2023physics3.2, wang2024grokked}. Comparison involves models evaluating two or more entities to make judgments, such as determining whether the attribute value of one entity is greater or smaller than that of another. Inverse search tests a model's ability to generate predictions in the reverse order of the training task. For example, this could involve applying the model to identify an entity based on its attributes when it was originally trained to predict the attributes of entities. Allen-Zhu et al. found that generative models struggle with inverse search unless pretrained for it specifically \citep{allenshu2023physics3.2}.

While models can recall individual facts well, they struggle with multi-hop reasoning that requires `chain-of-thought' logic. \citet{soheeyang2024multihopreasoning} found evidence of latent multi-hop reasoning for specific fact compositions, but noted it is highly contextual. Wang et al. demonstrated that Transformers can learn implicit reasoning, but only after extended training beyond typical overfitting thresholds \citep{wang2024grokked}. Composition tasks have also been studied in machine translation, testing whether models can generalize learned command components to new conjunctions, such as repeating actions \citep{lake2018generalization}.

\subsection{Models}\label{apd:first_models}

\noindent\textbf{Multi Layer Perceptrons (\MLP)} - A neural network architecture composed of stacked Fully Connected Neural Networks trained with backpropagation \citep{nair2010_mlp, fukushima1975_mlp, rosenblatt1958_mlp}. 

\noindent\textbf{Neural Hierarchical Interpolation for Time Series (\NHITS)} - A deep learning model that applies multi-rate input pooling, hierarchical interpolation, and backcast residual connections together to generate additive predictions with different signal bands \citep{challu_olivares2022_nhits}.

\noindent\textbf{Long Short-Term Memory Recurrent Neural Network (\LSTM)} - A recurrent neural network (\RNN) architecture that transforms hidden states from a multi-layer \LSTM\ encoder into contexts which are used to generate forecasts using \MLP s  \citep{sak2014_lstm}.

\noindent\textbf{Temporal Convolution Network (\TCN)} - A 1D causal-convolutional network architecture that transforms hidden states into contexts which are used as inputs to \MLP\ decoders to generate forecasts. Causal convolutions are used to generate contexts by convolving the prediction at time $t$ only with elements from time $t$ and earlier \citep{bai2018_tcn, oord2016_tcn}.

\noindent\textbf{TimesNet (\TimesNet)} - A deep learning architecture that transforms the original 1D time series into a set of
2D tensors based on multiple periods to capture intra- and inter-period variations modeled by 2D kernels \citep{wu2023timesnettemporal2dvariationmodeling}.

\noindent\textbf{VanillaTransformer (\VanillaTransformer)} - An encoder-decoder architecture with a multi-head attention mechanism that uses autoregressive features from a convolution network, window-relative positional embeddings from harmonic functions, and absolute positional embeddings from calendar data. An MLP decoder outputs time series predictions in a single pass \citep{vaswani_2021_attentionisallyouneed, zhou2021informerefficienttransformerlong}.

\noindent\textbf{Temporal Fusion Transformer (\TFT)} - An attention-based deep learning architecture that learns temporal relationships at different scales using \LSTM s for local processing and self-attention layers to model long-term dependencies. It also leverages variable selection networks and a series of gating layers to suppress unnecessary processing in the architecture \citep{lim2021_tft}. 

\noindent\textbf{Autoformer (\Autoformer)} - An encoder-decoder architecture with a multi-head attention mechanism that uses autoregressive features from a convolution network and absolute positional embeddings from calendar data. Decomposed trend and seasonal components are obtained using a moving average filter and an Auto-Correlation mechanism is used to identify periodic dependencies and aggregate similar sub-series \citep{wu_2021_autoformer, vaswani_2021_attentionisallyouneed}.

\noindent\textbf{Informer (\Informer)} - An encoder-decoder architecture with a multi-head attention mechanism that has three key features: a ProbSparse self-attention mechanism with \(O(L \log L)\) complexity, a self-attention distilling process, and an MLP decoder that outputs time-series predictions in a single pass. It uses autoregressive features from a convolution network, window-relative positional embeddings from harmonic functions, and absolute positional embeddings from calendar data \citep{zhou2021informerefficienttransformerlong, 
vaswani_2021_attentionisallyouneed}.

\noindent\textbf{iTransformer (\iTransformer)} - An attention-based deep learning architecture that applies attention and feed-forward networks to inverted dimensions by embedding time points into variate tokens. The attention mechanism capture multivariate correlations while the feed-forward network learns nonlinear representations for each token \citep{liu2024itransformerinvertedtransformerseffective}.

\noindent\textbf{Patch Time Series Transformer (\PatchTST)} - An encoder-only, multi-head attention-based architecture that separates input time series into sub-series level patches as input tokens. Each channel is dedicated to a single univariate time series, and all channels use the same embedding and Transformer weights. \cite{nie2023patchtst}

% \noindent\textbf{Patch Time Series Transformer with Decoder-only Architecture (\PatchDecoder)} 
% We modify \PatchTST to create a decoder-only model that leverages input time series patches, masked causal attention, and sequential autoregressive predictions during inference, similar to decoder-only language models \citep{brown2020gpt, liu2018decoderlm}. Fig.~\ref{fig:patch_model_architectures} (middle) illustrates the decoder-only architecture.

% \noindent\textbf{Patch Time Series Transformer with Encoder-Decoder Architecture (\PatchEncoderDecoder)} 
% We modify \PatchTST to create an encoder-decoder model that leverages input time series patches. The encoder architecture is equivalent to \PatchTST while the decoder leverages masked causal attention and cross-attention as with vanilla encoder-decoder transformer models \citep{zhou2021informerefficienttransformerlong}. The decoder also outputs sequential autoregressive predictions during inference. Fig.~\ref{fig:patch_model_architectures} (right) illustrates the encoder-decoder architecture.

\noindent\textbf{LagLlama (\LagLlama)} \citep{rasul2024lagllama} - A foundation model for univariate probabilistic time series forecasting based on a decoder-only Transformer architecture that uses lags as covariates \citep{rasul2024lagllama}. We use pretrained model weights from the HuggingFace library \citep{wolf_2020_hgtransformers}. We use a context length of 32, as this is the parameter value on which the model was trained. Although the model can handle other context lengths, which may improve forecasting performance.

\noindent\textbf{Chronos (\Chronos)} - A family of pretrained time series forecasting models that creates token sequences through scaling and quantization for input to a language model. Probabilistic forecasts are generated by sampling future trajectories from historical data \citep{ansari2024chronos}. We use the `t5-efficient-small' pretrained model weights from the HuggingFace library \citep{wolf_2020_hgtransformers}.

\noindent\textbf{MOMENT (\MOMENT)} - \MOMENT is a family of foundation models for time series that uses an attention-based architecture and processes input time series into fixed-length patches with each mapped to a D-dimensional embedding. During pretraining, patches are masked to minimize reconstruction error, and a separate linear projection is trained to adapt embeddings for long-horizon forecasting and other tasks \citep{goswami2024moment}. We use the `MOMENT-1-large' pretrained model weights from the HuggingFace library \citep{wolf_2020_hgtransformers}.

\noindent\textbf{TimesFM (\TimesFM)} - An attention-based decoder-only Transformer architecture that leverages input time series patching, masked causal attention, and sequential patch output predictions, with flexible output patch size \citep{das2024TimesFM}. We use the `timesfm-1.0-200m' pretrained model weights from the HuggingFace library \citep{wolf_2020_hgtransformers}.

% \begin{figure*}[ht]
% \centering
% \includegraphics[width=0.5\textwidth, trim=0 0 190 0, clip]{images/architectures1.pdf}
% \includegraphics[width=0.49\textwidth, trim=0 0 280 0, clip]{images/architectures2.pdf}
% \caption{Comparison of three transformer architectures for time series forecasting with time series patching. The encoder-only model (PatchTST) \citep{nie2023patchtst} patches the input time series, applies bidirectional attention, and directly outputs predictions for the defined forecast horizon \textbf{(left)}. The decoder-only architecture patches the input similar to the encoder-only architecture. However, it utilizes masked self-attention, and makes sequential predictions when the patch length is shorter than the forecast horizon \textbf{(middle)}. The encoder-decoder architecture patches the input time series, with the decoder using the final patch as the start token. Cross-attention is applied in the decoder and sequential predictions are generated when the patch length is less than the horizon \textbf{(right)}.}
% \label{fig:patch_model_architectures}
% \end{figure*}

% % Patch model comparison table
% \input{tables/patch_model_comparison_table}

\subsection{Synthetic Data Parameters}\label{apd:synthetic_data_parameters}
We leverage open-source code from \MOMENT ~\citep{goswami2024moment} to generate synthetic sinusoidal time series with varying trend, frequency, baseline, and amplitude. We use the following parameter values for all implicit reasoning tasks: $a \in [1, 32]$, $b \in [3, 32]$, $c \in [-32, 32]$. $m \in [1, 32]$ was used for addition, subtraction, and multiplication composition tasks, as well as for inverse search tasks. $m \in [-32, 32]$ was used for function composition and comparison tasks. For composition tasks, we generate $n = 30$ in-distribution (ID) series with evenly spaced parameter values. For comparison tasks, we generate $n = 300$ ID series with evenly spaced parameter values. For the inverse search task, we generate $n = 900$ ID addition composition aggregates using $n = 30$ trend and seasonality series, each with evenly spaced parameter values.

\subsection{Model Training and Parameters}\label{apd:first_hyperparameters}
Models were trained using \texttt{Adam} optimizer \citep{2017_adam_optimizer}. We train models with the following parameters outlined in Table~\ref{tab:hyperparameters} to ensure consistent evaluation across architectures. The training loss function used was Mean Squared Error (MSE), where $H$ refers to the forecast horizon, $t$ refers to the time point at which forecasts are generated, $y$ refers to the target signal values, and $\hat{y}$ refers to the model's predicted values:
$$MSE(y_{t+1, t+H}, \hat{y}_{t+1, t+H}) = \frac{1}{H}\sum_{i=t+1}^{t+H}(y_i - \hat{y}_i)^2.$$
The deep learning models were trained using an NVIDIA A100 Tensor Core GPU. 

\begin{table}[h!]
    \caption{Common hyperparameter search space}
    \centering
    \label{tab:hyperparameters}
    \begin{tabular}{l|l} 
    \toprule
      \textbf{Hyperparameter} & \textbf{Considered Values}\\
      \hline
      Input size & 200 \\
      Learning rate & 1e-3 \\
      Batch size & 4 \\
      Windows batch size & 256 \\
      Dropout & 0.0\\
      Training steps & 2000 \\
      Validation check steps & 50 \\
      %Early stop patience steps & 5 \\
      Early stop patience steps & 5 \\
      Random seed & 0 \\
      Patch Length & DiscreteRange(1, 50, 100, 150, 200)\\
      Hidden size & 128 \\
      Model layers & 3 \\
      \bottomrule
    \end{tabular}
  \centering
\end{table}

\subsection{Evaluation Metrics}\label{apd:first_metrics}
We use \textbf{Mean Absolute Error (MAE)} to evaluate model performance. Here, $H$ refers to the forecast horizon, $t$ refers to the time point at which forecasts are generated, $y$ refers to the target signal values, and $\hat{y}$ refers to the model's predicted values:\\
$$MAE(y_{t+1, t+H}, \hat{y}_{t+1, t+H}) = \frac{1}{H}\sum_{i=t+1}^{t+H}|y_i - \hat{y}_i|.$$

\subsection{Broader Impacts}\label{apd:broader_impacts}
This work aims to enhance our understanding of whether time series models generalize well by learning the underlying concepts of temporal dynamics or merely memorizing specific patterns seen during training. We find that certain models generalize effectively in systematically orchestrated out-of-distribution scenarios, suggesting underexplored reasoning capabilities beyond simple pattern memorization. Since this research is foundational and does not involve direct applications, we do not anticipate any negative societal impacts. Instead, our findings may provide valuable insights for developing more data- and computationally efficient deep learning architectures. Additionally, our results may help define the limitations of time series models, ensuring they are not applied in contexts where poor generalization performance is anticipated.

\subsection{Limitations}\label{apd:limitations}
This study has several limitations that should be acknowledged. First, our analysis relies on synthetic data, which may not fully capture the complexities and nuances present in real-world datasets. While synthetic data allows for controlled experimentation, it may limit the generalizability of our findings to broader applications. Second, we employed consistent hyperparameters across models to ensure a fair evaluation of performance; however, different combinations of hyperparameters may yield improved results. Future work will explore the impact of hyperparameter variations on model performance.

\subsection{Reproducibility Statement}
All models are open-source. Models trained on synthetic data can be found in the \texttt{Neuralforecast} library \citep{olivares2022library_neuralforecast}. Pretrained foundation models including, \LagLlama~\citep{rasul2024lagllama}, \Chronos~\citep{ansari2024chronos}, \TimesFM~\citep{das2024TimesFM}, and \MOMENT~\citep{goswami2024moment}, are open-source, with model weights available in the Hugging Face library \citep{wolf_2020_hgtransformers}. All models were trained and evaluated on a computing cluster consisting of 128 AMD EPYC 7502 CPUs, 503 GB of RAM, and 8 NVIDIA RTX A6000 GPUs each with 49 GiB RAM. The synthetic datasets used in our study will be released publicly with the full paper.\label{section:appendixA}
\newpage
\section{Supplemental Results}
\begin{table}[h!]
\caption{Mean Absolute Error (MAE) averaged across series (standard deviation) for the \textit{composition} implicit reasoning tasks: addition (add.), subtraction (sub.), multiplication (mult.), and sinusoidal function (function). MAE results for models trained on in-distribution (ID) time series and evaluated on out-of-distribution (OOD) series are presented in the `Task' column. We also include baseline (base.) MAE results for models that are both trained and evaluated on OOD data. The best results are highlighted in bold, and the second best results are highlighted in blue.}
\label{tab:composition_results_table}
\begin{tabular}{l|c c|c c|c c |c c }
\toprule
\multirow{3}{*}{\textbf{Model}} & \multicolumn{8}{c}{\textbf{MAE}}\\
\cline{2-9}
{} & \multicolumn{2}{c}{\textbf{Add.}} & \multicolumn{2}{c}{\textbf{Sub.}} & \multicolumn{2}{c}{\textbf{Mult.}} & \multicolumn{2}{c}{\textbf{Function}}\\
\cline{2-9}
{} & \textbf{Task} & \textbf{Base.} & \textbf{Task} & \textbf{Base.} & \textbf{Task} & \textbf{Base.} & \textbf{Task} & \textbf{Base.}\\
\hline
\multirow{2}{*}{\MLP} & 1.604 & \textcolor{blue}{0.386} & 2.637 & \textcolor{blue}{0.350} & \textcolor{blue}{2.097} & 1.507 & 10.099 & \textcolor{blue}{2.730} \\
                      & \small{(1.173)} & \small{(0.284)} & \small{(1.648)} & \small{(0.197)} & \small{(1.559)} & \small{(1.127)} & \small{(6.372)} & \small{(2.3)} \\
\hline
\multirow{2}{*}{\DLinear} & \textbf{0.689} & 0.743 & \textbf{0.935} & 0.715 & 10.522 & 8.792 & 21.155 & 10.653 \\
                      & \small{(0.349)} & \small{(0.333)} & \small{(0.452)} & \small{(0.333)} & \small{(7.945)} & \small{(5.284)} & \small{(23.965)} & \small{(6.783)} \\
\hline
\multirow{2}{*}{\NHITS} & 1.427 & \textbf{0.216} & 2.36 & \textbf{0.184} & \textbf{2.065} & \textbf{1.194} & \textbf{3.868} & \textbf{1.736} \\
                      & \small{(1.089)} & \small{(0.132)} & \small{(1.489)} & \small{(0.215)} & \small{(1.687)} & \small{(0.813)} & \small{(2.207)} & \small{(1.547)} \\
\hline
\multirow{2}{*}{\TSMixer} & 1.398 & 1.185 & 4.862 & 1.348 & 16.609 & 11.427 & 20.943 & 13.008 \\
                      & \small{(0.766)} & \small{(0.511)} & \small{(2.167)} & \small{(0.651)} & \small{(9.531)} & \small{(7.569)} & \small{(16.653)} & \small{(7.985)} \\
\hline
\multirow{2}{*}{\LSTM} & 7.546 & 2.973 & 8.992 & 5.216 & 9.831 & 5.641 & 11.353 & 5.095 \\
                      & \small{(4.288)} & \small{(1.793)} & \small{(4.954)} & \small{(2.904)} & \small{(5.529)} & \small{(3.714)} & \small{(5.12)} & \small{(2.928)} \\
\hline
\multirow{2}{*}{\TCN} & 5.716 & 2.538 & 5.063 & 3.057 & 8.359 & 2.606 & 11.413 & 4.123 \\
                      & \small{(3.228)} & \small{(1.432)} & \small{(2.597)} & \small{(1.667)} & \small{(4.816)} & \small{(1.995)} & \small{(4.781)} & \small{(2.391)} \\
\hline
\multirow{2}{*}{\TimesNet} & 1.630 & 0.555 & 2.937 & 0.512 & 10.635 & 2.082 & 11.643 & 3.744 \\
                      & \small{(0.627)} & \small{(0.238)} & \small{(1.519)} & \small{(0.239)} & \small{(7.88)} & \small{(1.692)} & \small{(8.352)} & \small{(3.076)} \\
\hline
\multirow{2}{*}{\VanillaTransformer} & 2.432 & 0.721 & 2.993 & 0.658 & 8.249 & 5.032 & 11.291 & 7.341 \\
                      & \small{(1.468)} & \small{(0.228)} & \small{(1.69)} & \small{(0.152)} & \small{(4.939)} & \small{(3.492)} & \small{(7.334)} & \small{(4.552)} \\
\hline
\multirow{2}{*}{\TFT} & \textcolor{blue}{1.024} & 0.709 & 3.243 & 1.057 & 13.395 & 5.121 & 10.159 & 6.548 \\
                      & \small{(0.408)} & \small{(0.295)} & \small{(1.507)} & \small{(0.504)} & \small{(9.688)} & \small{(3.834)} & \small{(5.228)} & \small{(4.455)} \\
\hline
\multirow{2}{*}{\Autoformer} & 1.708 & 0.914 & 3.999 & 1.188 & 12.845 & 9.69 & 13.383 & 11.344 \\
                      & \small{(0.952)} & \small{(0.315)} & \small{(1.886)} & \small{(0.52)} & \small{(7.764)} & \small{(5.716)} & \small{(8.018)} & \small{(6.846)} \\
\hline
\multirow{2}{*}{\Informer} & 2.045 & 0.741 & 3.025 & 0.78 & 9.246 & 5.423 & 12.268 & 10.335 \\
                      & \small{(1.273)} & \small{(0.235)} & \small{(1.702)} & \small{(0.28)} & \small{(5.303)} & \small{(3.4)} & \small{(8.89)} & \small{(5.724)} \\
\hline
\multirow{2}{*}{\iTransformer} & 1.358 & 0.933 & 2.994 & 0.947 & 15.197 & 14.618 & 31.632 & 11.773 \\
                      & \small{(0.886)} & \small{(0.384)} & \small{(1.403)} & \small{(0.447)} & \small{(10.35)} & \small{(8.732)} & \small{(16.823)} & \small{(6.882)} \\
\hline
    \multirow{2}{*}{\PatchTST} & 1.717 & 0.435 & \textcolor{blue}{2.168} & 0.584 & 3.497 & \textcolor{blue}{1.600} & \textcolor{blue}{6.209} & 7.231 \\
                      & \small{(1.113)} & \small{(0.285)} & \small{(1.185)} & \small{(0.404)} & \small{(2.442)} & \small{(1.108)} & \small{(4.009)} & \small{(6.031)} \\
\hline
% \multirow{2}{*}{\PatchDecoder} & x & x & x & x & x & x & x & x \\
%                       & \small{(x)} & \small{(x)} & \small{(x)} & \small{(x)} & \small{(x)} & \small{(x)} & \small{(x)} & \small{(x)} \\
% \hline
% \multirow{2}{*}{\PatchEncoderDecoder} & x & x & x & x & x & x & x & x \\
%                       & \small{(x)} & \small{(x)} & \small{(x)} & \small{(x)} & \small{(x)} & \small{(x)} & \small{(x)} & \small{(x)} \\
% \hline
\hline
\textbf{TSFMs} & \multicolumn{8}{c}{\textbf{Zero-shot Forecasting}} \\
\hline
\multirow{2}{*}{\LagLlama} & \multicolumn{2}{c}{5.181} & \multicolumn{2}{c}{4.263} & \multicolumn{2}{c}{9.904} & \multicolumn{2}{c}{17.684} \\
                      & \multicolumn{2}{c}{\small{(2.720)}} & \multicolumn{2}{c}{\small{(2.475)}} & \multicolumn{2}{c}{\small{(5.795)}} & \multicolumn{2}{c}{\small{(7.206)}} \\
\hline
\multirow{2}{*}{\Chronos} & \multicolumn{2}{c}{1.064} & \multicolumn{2}{c}{1.124} & \multicolumn{2}{c}{1.922} & \multicolumn{2}{c}{2.332} \\
                      & \multicolumn{2}{c}{\small{(1.133)}} & \multicolumn{2}{c}{\small{(1.228)}} & \multicolumn{2}{c}{\small{(2.499)}} & \multicolumn{2}{c}{\small{(2.596)}} \\
\hline
\multirow{2}{*}{\TimesFM} & \multicolumn{2}{c}{0.475} & \multicolumn{2}{c}{0.467} & \multicolumn{2}{c}{0.355} & \multicolumn{2}{c}{0.672} \\
                      & \multicolumn{2}{c}{\small{(0.375)}} & \multicolumn{2}{c}{\small{(0.457)}} & \multicolumn{2}{c}{\small{(0.336)}} & \multicolumn{2}{c}{\small{(0.655)}} \\
\hline
\multirow{2}{*}{\MOMENT} & \multicolumn{2}{c}{2.428} & \multicolumn{2}{c}{3.140} & \multicolumn{2}{c}{9.960} & \multicolumn{2}{c}{10.877} \\
                      & \multicolumn{2}{c}{\small{(1.392)}} & \multicolumn{2}{c}{\small{(1.564)}} & \multicolumn{2}{c}{\small{(6.022)}} & \multicolumn{2}{c}{\small{(6.347)}} \\
\bottomrule
\end{tabular}
\end{table}

\begin{table}[h!]
\centering
\caption{Model forecast Mean Absolute Error (MAE) averaged across series (standard deviation). MAE results for models trained on in-distribution (ID) time series and evaluated on out-of-distribution (OOD) series are presented in the `Task' column. We also include baseline (base.) MAE results for models that are both trained and evaluated on OOD data. The best results are highlighted in bold, and the second best results are highlighted in blue.}
%\label{tab:comparison_invsearch_table}
\begin{minipage}{.48\linewidth}
\centering
\subcaption{\textit{Comparison} Task}
\label{tab:comparison_results_table}
\begin{tabular}{l|c c}
\toprule
\multirow{3}{*}{\textbf{Model}} & \multicolumn{2}{c}{\textbf{MAE}}\\
\cline{2-3}
{} & \multicolumn{2}{c}{\textbf{Comparison}} \\
\cline{2-3}
\textbf{} & \textbf{Task} & \textbf{Base.} \\
\hline
\multirow{2}{*}{\MLP}  & 2.492 & 1.945\\
                      & \small{(4.831)} & \small{(4.138)} \\
\hline
\multirow{2}{*}{\DLinear} & \textbf{1.072} & \textcolor{blue}{1.342}\\
                      & \small{(1.173)} & \small{(1.848)} \\
\hline
\multirow{2}{*}{\NHITS}  & 2.218 & 1.750\\
                      & \small{(4.113)} & \small{(3.749)}  \\
\hline
\multirow{2}{*}{\TSMixer} & 2.488 & 2.391 \\
                       & \small{(3.371)} & \small{(3.570)}\\
\hline
\multirow{2}{*}{\LSTM}  & 3.174 & 4.663\\
                      & \small{(3.767)} & \small{(5.571)}  \\
\hline
\multirow{2}{*}{\TCN}  & 2.467 & 2.913 \\
                      & \small{(3.002)}  & \small{(3.591)}\\
\hline
\multirow{2}{*}{\TimesNet} & 3.570 & 1.972\\
                       & \small{(6.309)} & \small{(4.001)}\\
\hline
\multirow{2}{*}{\VanillaTransformer} & 2.120 & 2.040 \\
                       & \small{(4.071)} & \small{(4.066)}\\
\hline
\multirow{2}{*}{\TFT} & 2.045 & 1.986 \\
                       & \small{(3.802)} & \small{(3.864)}\\
\hline
\multirow{2}{*}{\Autoformer}  & 3.966  & 3.785\\
                       & \small{(5.097)} & \small{(5.145)}\\
\hline
\multirow{2}{*}{\Informer}  & 2.131 & 1.999 \\
                       & \small{(3.805)} & \small{(3.674)}\\
\hline
\multirow{2}{*}{\iTransformer}  & 2.885  & 3.106\\
                      & \small{(4.052)}  & \small{(4.429)} \\
\hline
\multirow{2}{*}{\PatchTST}  & \textcolor{blue}{1.740}  & \textbf{1.204} \\
                       & \small{(3.084)} & \small{(2.678)}\\
\hline
% \multirow{2}{*}{\PatchDecoder} & x & x \\
%                       & \small{(x)} & \small{(x)} \\
% \hline
% \multirow{2}{*}{\PatchEncoderDecoder} & x & x \\
%                       & \small{(x)} & \small{(x)} \\
%\hline
\hline
\textbf{TSFMs} & \multicolumn{2}{c}{\textbf{Zero-Shot}} \\
\hline
\multirow{2}{*}{\LagLlama} & \multicolumn{2}{c}{5.307} \\
                      & \multicolumn{2}{c}{\small{(7.512)}} \\
\hline
\multirow{2}{*}{\Chronos} & \multicolumn{2}{c}{0.704} \\
                      & \multicolumn{2}{c}{\small{(1.362)}} \\
\hline
\multirow{2}{*}{\TimesFM} & \multicolumn{2}{c}{0.413} \\
                      & \multicolumn{2}{c}{\small{(0.775)}} \\
\hline
\multirow{2}{*}{\MOMENT} & \multicolumn{2}{c}{3.381} \\
                      & \multicolumn{2}{c}{\small{(4.557)}} \\
\bottomrule
\end{tabular}
\end{minipage}%
\hspace{0.02\linewidth} % Add space between subtables
\begin{minipage}{.48\linewidth}
\centering
\subcaption{\textit{Inverse Search} Task}
\label{tab:inverse_search_results_table}
\begin{tabular}{l|c c}
\toprule
\multirow{3}{*}{\textbf{Model}} & \multicolumn{2}{c}{\textbf{MAE}}\\
\cline{2-3}
{} & \multicolumn{2}{c}{\textbf{Inverse Search}} \\
\cline{2-3}
\textbf{} & \textbf{Task} & \textbf{Base.} \\
\hline
\multirow{2}{*}{\MLP} & 0.632 & \textcolor{blue}{0.076} \\
                      & \small{(0.523)} & \small{(0.040)} \\
\hline
\multirow{2}{*}{\DLinear} & 0.539 & 0.444 \\
                      & \small{(0.433)} & \small{(0.389)} \\
\hline
\multirow{2}{*}{\NHITS} & \textbf{0.214} & \textbf{0.035} \\
                      & \small{(0.231)} & \small{(0.055)} \\
\hline
\multirow{2}{*}{\TSMixer} & 1.622 & 1.431 \\
                      & \small{(0.697)} & \small{(0.647)} \\
\hline
\multirow{2}{*}{\LSTM} & 1.784 & 4.161 \\
                      & \small{(1.617)} & \small{(4.665)} \\
\hline
\multirow{2}{*}{\TCN} & 1.666 & 3.113 \\
                      & \small{(1.444)} & \small{(3.504)} \\
\hline
\multirow{2}{*}{\TimesNet} & 0.646 & 0.438 \\
                      & \small{(0.384)} & \small{(0.385)} \\
\hline
\multirow{2}{*}{\VanillaTransformer} & 0.551 & 0.279 \\
                      & \small{(0.531)} & \small{(0.248)} \\
\hline
\multirow{2}{*}{\TFT} & 0.574 & 0.961 \\
                      & \small{(0.464)} & \small{(0.504)} \\
\hline
\multirow{2}{*}{\Autoformer} & 1.180 & 0.804  \\
                      & \small{(0.527)} & \small{(0.335)} \\
\hline
\multirow{2}{*}{\Informer} & 0.877 & 0.436  \\
                      & \small{(0.425)} & \small{(0.210)} \\
\hline
\multirow{2}{*}{\iTransformer} & 1.164 & 1.143  \\
                      & \small{(0.507)} & \small{(0.560)}  \\
\hline
\multirow{2}{*}{\PatchTST} & \textcolor{blue}{0.394} & 0.293  \\
                      & \small{(0.545)} & \small{(0.195)} \\
\hline
% \multirow{2}{*}{\PatchDecoder} & x & x \\
%                       & \small{(x)} & \small{(x)} \\
% \hline
% \multirow{2}{*}{\PatchEncoderDecoder} & x & x \\
%                       & \small{(x)} & \small{(x)} \\
% \hline
\hline
\textbf{TSFMs} & \multicolumn{2}{c}{\textbf{Zero-Shot}} \\
\hline
\multirow{2}{*}{\LagLlama} & \multicolumn{2}{c}{2.656} \\
                      & \multicolumn{2}{c}{\small{(2.724)}} \\
\hline
\multirow{2}{*}{\Chronos} & \multicolumn{2}{c}{0.096} \\
                      & \multicolumn{2}{c}{\small{(0.088)}} \\
\hline
\multirow{2}{*}{\TimesFM} & \multicolumn{2}{c}{0.070} \\
                      & \multicolumn{2}{c}{\small{(0.075)}}  \\
\hline
\multirow{2}{*}{\MOMENT} & \multicolumn{2}{c}{1.502} \\
                      & \multicolumn{2}{c}{\small{(1.225)}} \\
\bottomrule
\end{tabular}
\end{minipage}
\end{table}

\begin{figure}[h!]
     \centering
     \begin{subfigure}[b]{0.495\textwidth}
         \centering
         \includegraphics[width=\textwidth]{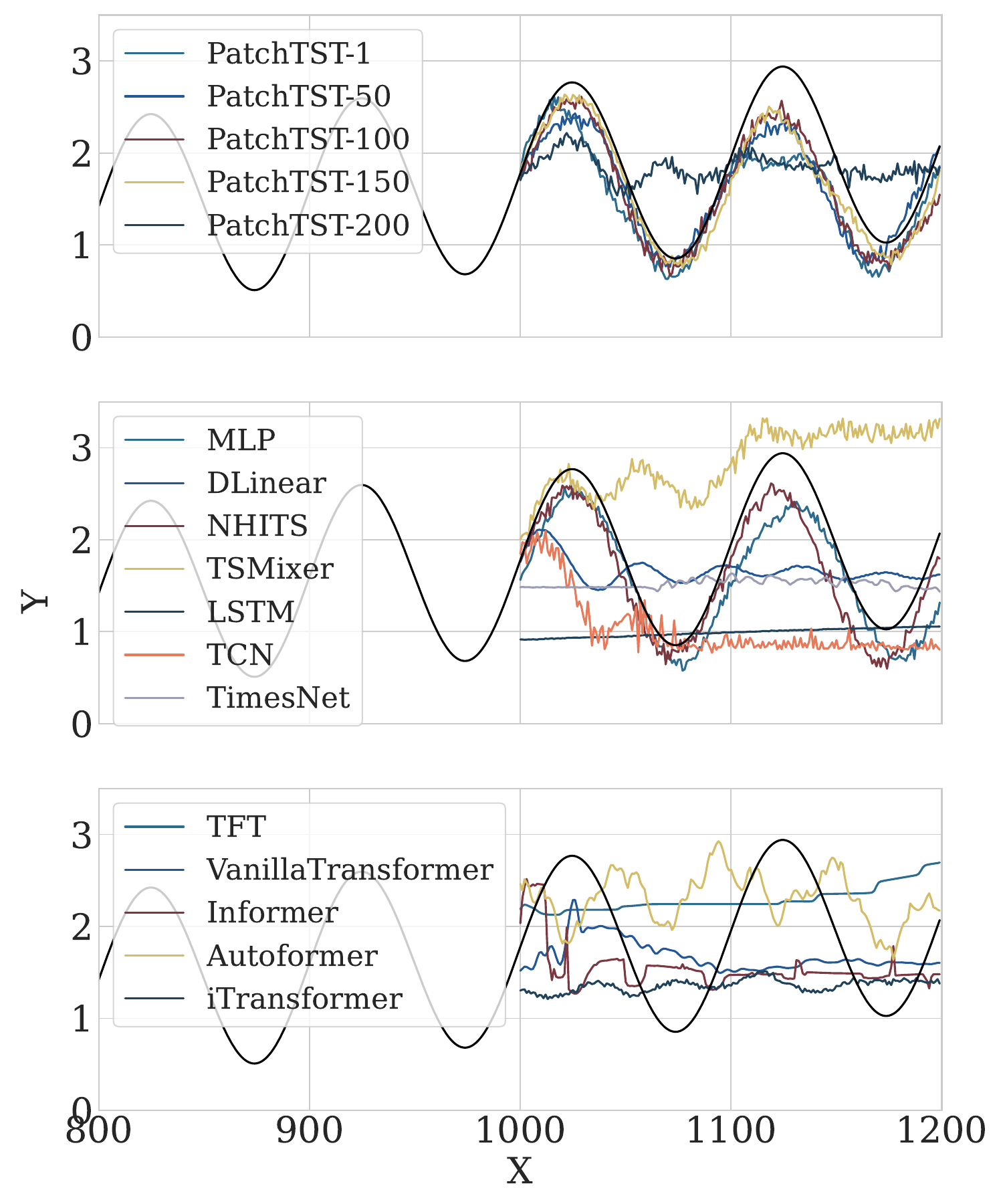}
         \caption{Addition}
         \label{fig:addition_forecast_examples}
     \end{subfigure}
     \begin{subfigure}[b]{0.495\textwidth}
         \centering
         \includegraphics[width=\textwidth]{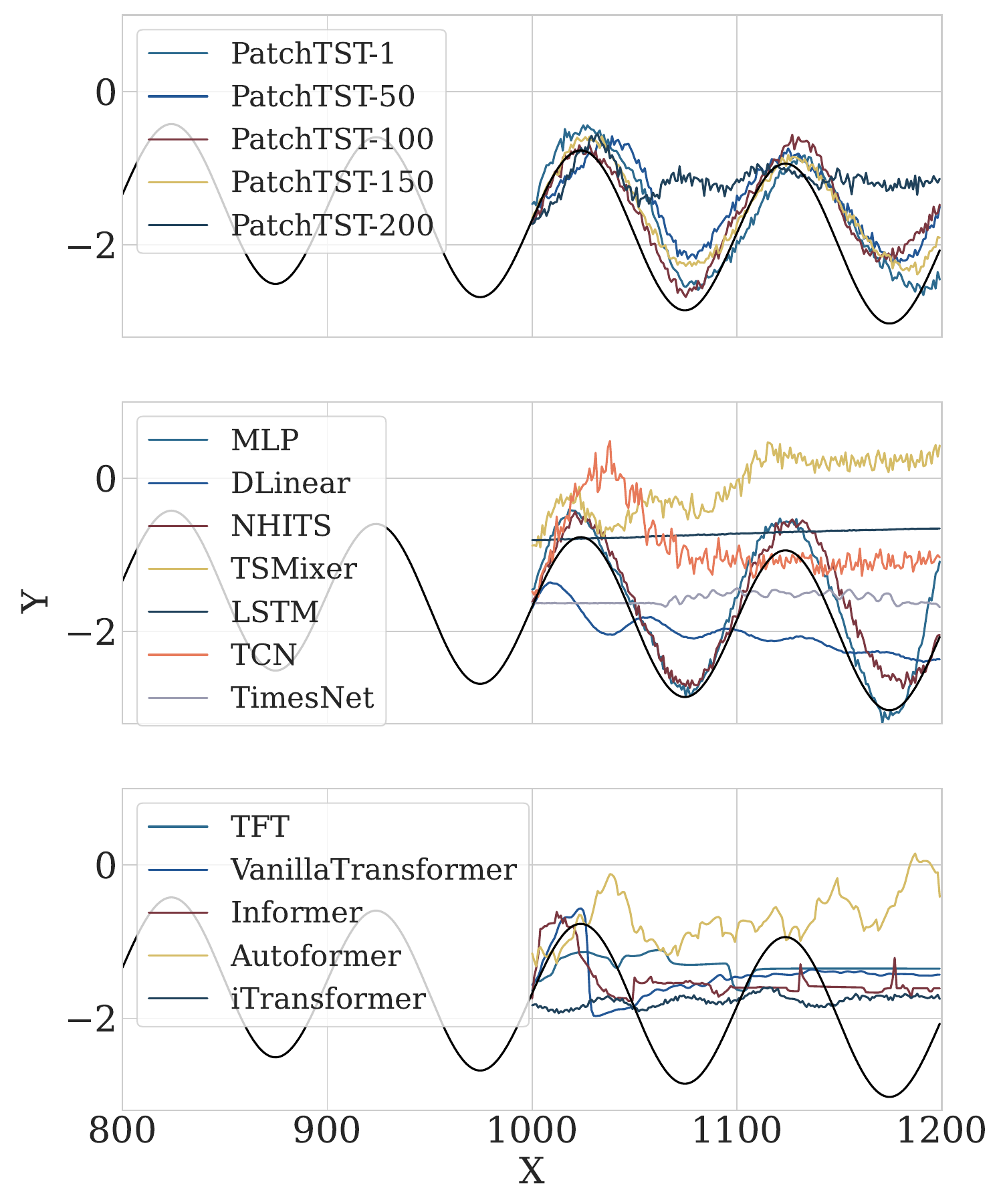}
         \caption{Subtraction}
         \label{fig:subtraction_forecast_examples}
     \end{subfigure}
        \caption{Model time series forecasts for addition (a) and subtraction (b) \textit{composition} tasks.}
        \label{fig:compositon_forecast_examples}
\end{figure}

\begin{figure}
     \centering
     \includegraphics[width=\textwidth]{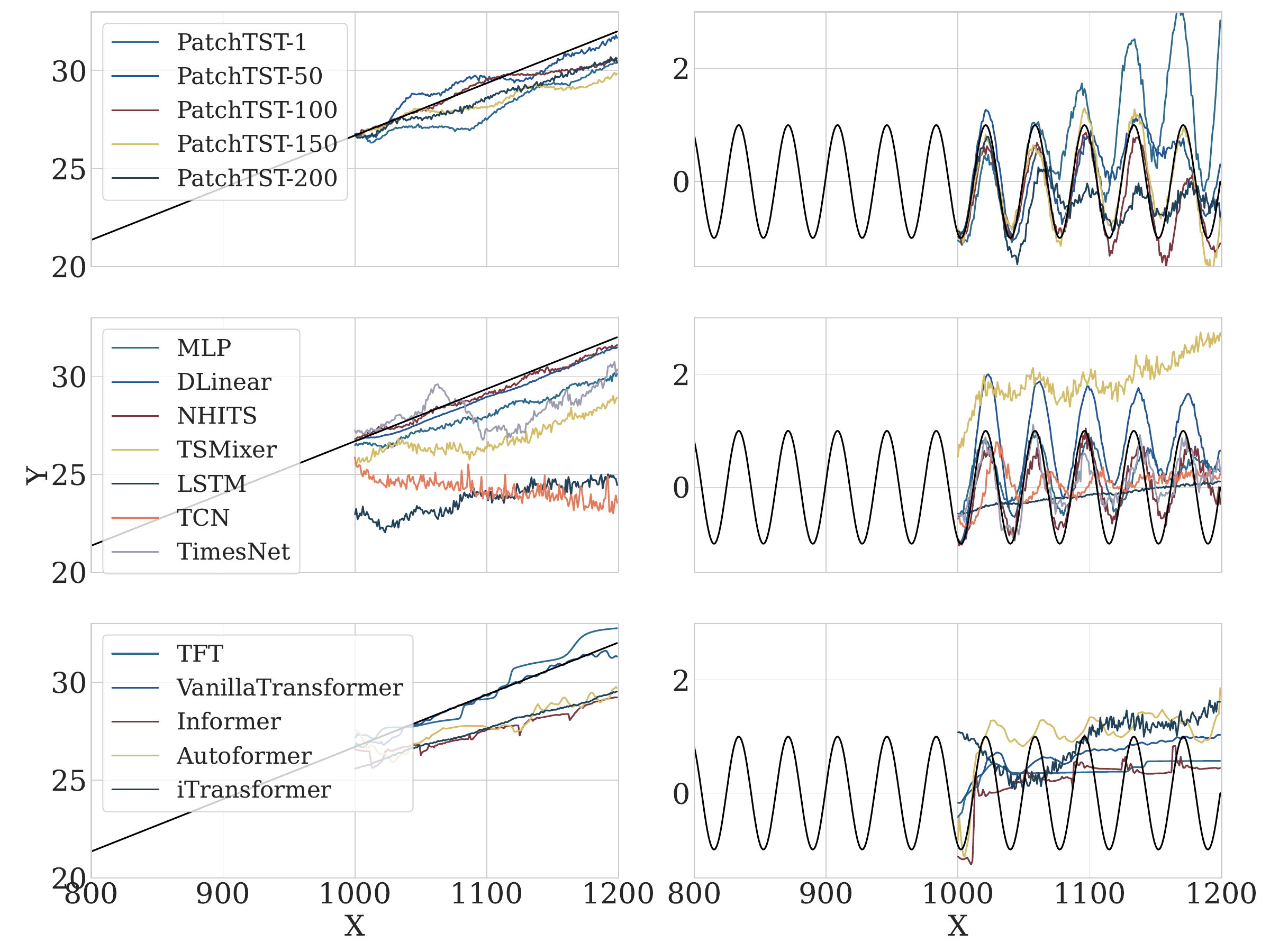}
     \caption{Model time series forecasts for the \textit{inverse search} task.}
     \label{fig:inverse_search_forecast_examples}
\end{figure} \label{section:appendixB}

\end{document}